%% file: neurips_2026.tex
\documentclass{article}
\pdftrailerid{}
\PassOptionsToPackage{numbers, compress}{natbib}

\usepackage[main,final]{neurips_2026}

\usepackage[utf8]{inputenc} 
\usepackage[T1]{fontenc}    
\usepackage{booktabs}       
\usepackage{amsfonts}       
\usepackage{amsmath}        
\usepackage{amsthm}         
\usepackage{mathtools}      
\usepackage{algorithm}      
\usepackage{algpseudocode}  
\usepackage{hyperref}       
\hypersetup{pdfcreator={},pdfproducer={}}
\usepackage{url}            
\usepackage{nicefrac}       
\usepackage{microtype}      
\usepackage{xcolor}         
\usepackage{etoolbox}
\newcommand{\blackbibitem}[2][]{%
  \color{black}%
  \originalbibitem[#1]{#2}}
\AtBeginEnvironment{thebibliography}{\let\originalbibitem\bibitem\let\bibitem\blackbibitem}
\usepackage{colortbl}       
\usepackage{multirow}       
\usepackage{graphicx}       
\usepackage{subcaption}     
\usepackage{wrapfig}        
\usepackage[most]{tcolorbox}
\usepackage{fvextra}         

\theoremstyle{definition}

\title{Negative-Only Policy Optimization for \\ One-Sided Verifiable Rewards}

\author{%
  Jiacheng Xu\textsuperscript{1,2} \quad Shuo He\textsuperscript{1} \quad Fuxiang Zhang\textsuperscript{1,2} \quad Chaojie Wang\textsuperscript{2} \quad Bo An\textsuperscript{1} \\
  \textsuperscript{1}Nanyang Technological University, Singapore \quad \textsuperscript{2}Skywork AI \\
  \texttt{jiacheng005@e.ntu.edu.sg}, \texttt{boan@ntu.edu.sg}
}

\begin{document}

\maketitle

\input{main}

\bibliographystyle{plainnat}
\bibliography{references}

\appendix

\input{appendix}

\end{document}

%% file: main.tex
\begin{abstract}
Reinforcement learning with verifiable rewards (RLVR) typically relies on a verifier that can reliably judge whether sampled solutions are correct.
However, in many realistic settings, complete verification is generally unavailable, while cheap checks can only falsify failures.
We formalize this regime as \emph{one-sided verifiability} (OSV): verified negatives are reliable counterexamples, but verified positives are merely non-falsified candidates and may contain many false positives.
This asymmetry makes standard RLVR unreliable, because reinforcing verified positives can directly reward shortcuts that pass partial checks.
To address this issue, we propose \emph{Negative-Only Policy Optimization} (NOPO), which treats verified positives as unlabeled and updates the policy only by suppressing trusted verified negatives.
Specifically, NOPO combines a reference-relative negative suppression loss with a group-level adapter that scales each update by the proportion of non-falsified responses.
Across unit-test, constraint-check, and LLM self-verifiers, NOPO improves $\text{pass}@k$ over baselines on code, logic, and math benchmarks when oracle rewards are unavailable.
These gains persist under inference-time scaling and cross-benchmark transfer.
\end{abstract}

\begin{center}
\includegraphics[width=\textwidth]{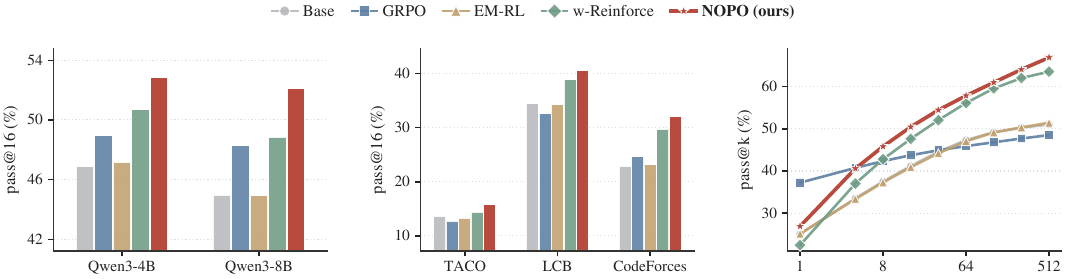}
\captionsetup{hypcap=false}
\captionof{figure}{Overview of NOPO results.
Left: average $\text{pass}@16$ across benchmarks from the main results in Table~\ref{tab:main-result}.
Middle: cross-benchmark transfer of the Qwen3-4B model trained on CodeContests to other code benchmarks.
Right: $\text{pass}@k$ scaling on CodeContests for the Qwen3-4B model.}
\label{fig:teaser}
\end{center}

\input{content/intro}
\input{content/background}
\input{content/osv}
\input{content/nopo}
\input{content/experiments}
\input{content/related}
\input{content/conclusion}

%% file: content/intro.tex
\section{Introduction}
\label{sec:intro}

\begin{figure*}[t]
  \vskip -0.1in
  \centering
  \includegraphics[width=\textwidth]{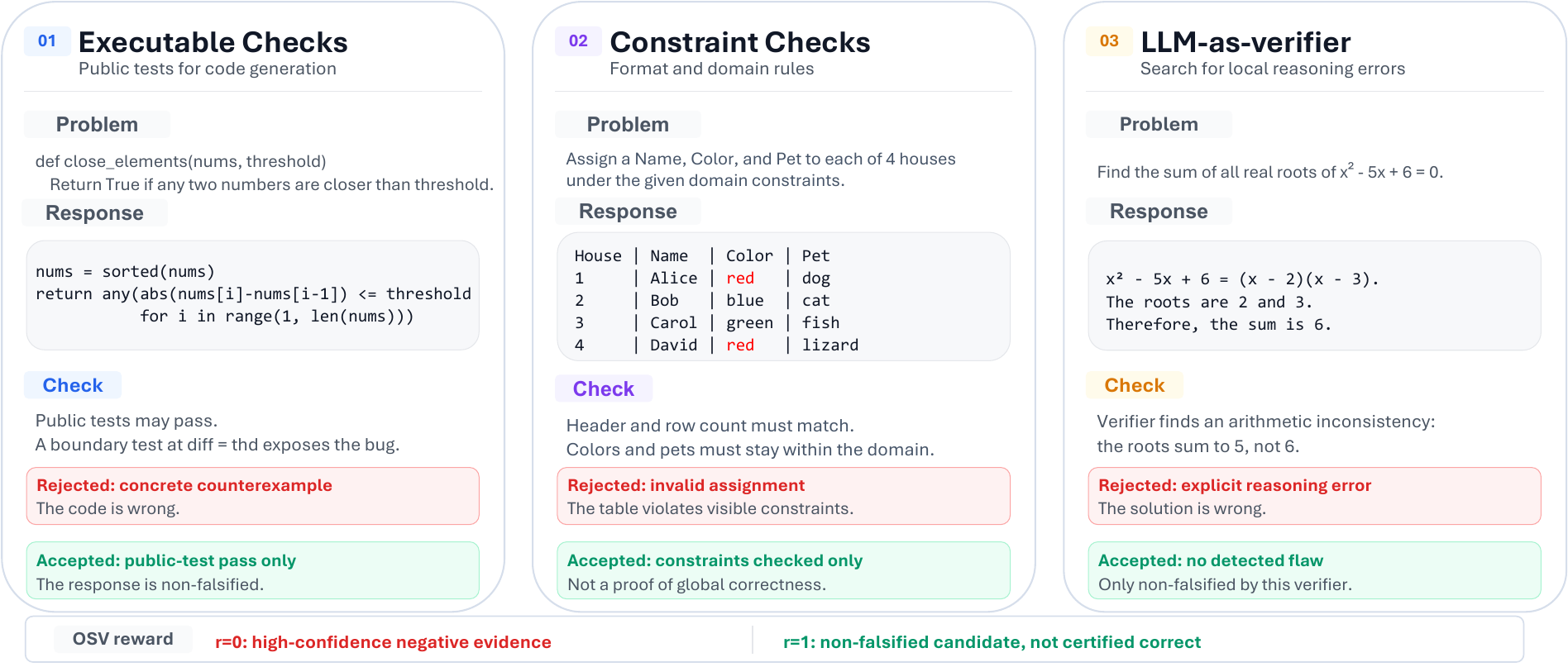}
  \caption{Illustration of one-sided verifiability across executable checks, constraint checks, and LLM-as-verifier settings. A rejection provides concrete evidence of failure, whereas acceptance only means that no checked violation was found; it does not certify correctness.}
  \label{fig:osv}
  \vskip -0.15in
\end{figure*}

Reinforcement learning with verifiable rewards (RLVR) has become a central recipe for improving reasoning-oriented LLMs, using a verifier to score sampled solutions and optimize the policy based on the resulting scores~\citep{tulu3,deepseekr1,openai_o1,kimi_k15}.
This paradigm is most straightforward when the reward is complete: the verifier can reliably determine whether each sampled solution is correct.
Outside curated training environments, however, constructing such complete verifiers becomes a bottleneck.
Users may specify a task without being able to provide a full checker, and in open-ended scientific or reasoning domains, certifying a proposed answer can be as difficult as producing one.
The central question is therefore not only how to learn from verifiable rewards, but how to learn when verification itself is incomplete.

The key observation of this paper is that incomplete verification is often asymmetric.
Complete correctness may be unavailable, expensive, or as hard as solving the task, while local failures are much easier to detect.
For example, in code generation, a failing unit test exposes a concrete bug even though a passing test suite does not establish full program correctness~\citep{humaneval,evalplus,xu2026tcs}.
In structured reasoning tasks, explicit format or domain constraints can reject invalid answers without solving the task~\citep{zebralogic}.
When no reliable task-specific checker is available, an LLM verifier may still identify obvious flaws in a proposed solution~\citep{cobbe2021verifiers,mtbench,genrm2024}.
As illustrated in Figure~\ref{fig:osv}, these checks provide one-sided evidence: a rejection is high-confidence evidence of failure, whereas an acceptance only means that no checked violation was found.
We study this regime as \emph{one-sided verifiability} (OSV): verifier-negative judgments have high precision, while verifier-positive judgments may contain a non-negligible false-positive component.
Under OSV, a reward of $0$ is high-confidence negative evidence, but a reward of $1$ is not a trusted positive label and only a non-falsified candidate.

Existing approaches do not directly address this regime.
Verifier-free or self-training methods construct surrogate learning signals from majority agreement, novelty, or confidence-like statistics~\citep{ttrl2025,evolrl2025,emrl2025,xu2026erpo}.
These methods show that useful signals can be obtained without ground-truth labels, but they do not exploit a common form of external feedback: reliable rejection without reliable acceptance.
Conversely, standard RLVR methods such as GRPO~\citep{grpo} use the available verifier as a binary reward, reinforcing verifier-positive samples and suppressing verifier-negative samples.
Under OSV, this binary interpretation is misleading.
False positives are not merely noisy evaluations; once used as rewards, they become attractors of the policy optimization dynamics.
Shortcut solutions that exploit unchecked failure modes receive direct positive gradients whenever they pass the partial verifier, allowing the policy to improve verifier pass rate without improving true correctness~\citep{skalse2022,darkrich2024}.

In this paper, we propose \emph{Negative-Only Policy Optimization} (NOPO), a policy optimization method designed for one-sided verifiable rewards.
NOPO uses the verifier only for the evidence it can safely provide: high-confidence negatives.
It suppresses verifier-negative trajectories while treating verifier-positive trajectories as unlabeled rather than correct.
This negative-only principle creates a new optimization challenge: suppressing failures redistributes probability mass, but an all-failed rollout group provides no grounded direction for where that mass should move.
NOPO addresses this challenge with two safeguards.
At the sample level, it uses a reference-relative NPO loss whose gradient self-attenuates once a verifier-negative response is sufficiently downweighted relative to the reference policy.
At the group level, it scales negative updates by the amount of non-falsified support in the same rollout group.
Importantly, this support weight does not treat verifier-positive samples as correct demonstrations; it uses only their aggregate count to indicate whether the current policy has produced non-falsified alternatives toward which probability mass can be conservatively redistributed.
Together, these components enable policy improvement from reliable negative evidence without directly reinforcing noisy positives.

Our main contributions are:
{\setlength{\leftmargini}{2em}
\begin{enumerate}
    \item \textbf{One-sided verifiability.}
    We identify and formalize a common regime in realistic LLM tasks where local falsification is reliable but complete verification is unavailable.
    Under OSV, verifier negatives provide high-confidence evidence of failure, whereas verifier positives are only non-falsified candidates rather than trusted positive labels.

    \item \textbf{Negative-Only Policy Optimization.}
    We propose NOPO, an RL method that learns from high-confidence negatives without directly reinforcing noisy positives.
    NOPO combines sample-level reference-relative NPO suppression with a group-level non-falsified support weight, yielding conservative policy improvement under one-sided verifiable rewards.

    \item \textbf{Empirical validation.}
    We validate the OSV noise profile across multiple verifier types, showing that negative judgments are substantially cleaner than positive judgments.
    Under OSV, NOPO improves the model's capability frontier over standard RLVR baselines, and the gains persist under inference-time scaling and cross-benchmark transfer.
\end{enumerate}
}

%% file: content/background.tex
\section{Preliminaries}
\label{sec:preliminaries}

\paragraph{Reinforcement Learning with Verifiable Rewards.}

Reinforcement Learning with Verifiable Rewards (RLVR) optimizes a policy $\pi_\theta$ using reward signals produced by a verifier.
Given an input $x$ and a sampled output $y \sim \pi_{\theta_{\mathrm{old}}}(\cdot \mid x)$, the verifier returns a reward $r(x,y)$.
A generic policy-gradient-style RLVR objective can be written as
\begin{equation}
\label{eq:rlvr}
\resizebox{0.92\linewidth}{!}{$\displaystyle \mathcal{J}_{\mathrm{RLVR}}(\theta)=\mathop{\mathbb{E}}\limits_{x\sim\mathcal{D}}\,\mathop{\mathbb{E}}\limits_{y\sim\pi_{\theta_{\mathrm{old}}}(\cdot\mid x)}\Big[\min\!\big(\rho(x,y)A(x,y),\,\bar{\rho}(x,y)A(x,y)\big)-\lambda_{\mathrm{KL}}\,D_{\mathrm{KL}}(\pi_\theta\,\|\,\pi_{\mathrm{ref}})\Big].$}
\end{equation}
Here $\rho(x,y) \coloneqq \frac{\pi_\theta(y\mid x)}{\pi_{\theta_{\mathrm{old}}}(y\mid x)}$ is the importance ratio, $\bar{\rho}(x,y) \coloneqq \operatorname{clip}(\rho(x,y), 1-\epsilon, 1+\epsilon)$ is its clipped version, $A(x,y)$ is an advantage term constructed from verifier rewards, and $\lambda_{\mathrm{KL}}$ controls KL regularization toward a reference policy $\pi_{\mathrm{ref}}$.
Different RLVR algorithms mainly differ in how they construct $A(x,y)$ from sampled rollouts and verifier signals.

GRPO~\citep{grpo} is one concrete RLVR instantiation and serves as a baseline in our experiments as well as the reference point for our later analysis.
For each prompt $x$, GRPO samples a group of $G$ outputs $\{y_i\}_{i=1}^G$ and defines
\begin{equation}
\label{eq:grpo}
A_i \coloneqq \frac{r_i - \operatorname{mean}(r_1,\dots,r_G)}{\operatorname{std}(r_1,\dots,r_G)},
\end{equation}
where $r_i$ is the verifier reward of response $y_i$.

\paragraph{Negative Preference Optimization.}
\label{sec:npo-prelim}

LLM unlearning studies how to reduce a model's likelihood on a specified set of undesirable examples or behaviors while preserving its general capabilities.
This is related to our setting in a limited but important sense: both require lowering the likelihood of trusted negative outputs without pushing the model arbitrarily far from its original behavior.
Negative Preference Optimization (NPO)~\citep{npo2024} is a representative unlearning objective for this purpose.
For a negative response $y$ to prompt $x$, define
\[
\Delta_\theta(x,y)
\coloneqq
\log \pi_\theta(y\mid x)-\log \pi_{\mathrm{ref}}(y\mid x).
\]
The per-example NPO loss is
\begin{equation}
\label{eq:npo-loss}
\mathcal{L}_{\mathrm{NPO},\beta}(\theta;x,y)
=
\frac{2}{\beta}
\log\!\left(1+\exp\!\left(\beta\Delta_\theta(x,y)\right)\right)
\qquad \beta>0 .
\end{equation}
A training objective averages this term over the trusted negative set.
The inverse temperature $\beta$ controls how sharply the loss transitions from penalizing likely negatives to attenuating already-suppressed negatives.
The loss penalizes a response relative to its reference probability without requiring an explicit preferred response.
Its derivative with respect to $\log \pi_\theta(y\mid x)$ is $2\sigma(\beta\Delta_\theta(x,y))$.
Thus, the penalty is large when the current policy still assigns the negative response probability comparable to or above the reference policy, and becomes small once the response has already been suppressed below the reference.

%% file: content/osv.tex
\section{One-Sided Verifiability}
\label{sec:osv}

\begin{table}[t]
\centering
\small
\caption{Pre-RL verifier noise across main benchmarks.}
\label{tab:osv}
\resizebox{\linewidth}{!}{
\begin{tabular}{ll cccc cccc}
\toprule
 & & \multicolumn{4}{c}{\textbf{Qwen3-4B}} & \multicolumn{4}{c}{\textbf{Qwen3-8B}} \\
\cmidrule(lr){3-6} \cmidrule(lr){7-10}
\textbf{Benchmark} & \textbf{Verifier} & pass@16 & $q_{\mathrm{neg}}$ & $q_{\mathrm{pos}}$ & $r_{\mathrm{neg}}$ & pass@16 & $q_{\mathrm{neg}}$ & $q_{\mathrm{pos}}$ & $r_{\mathrm{neg}}$ \\
\midrule
TACO test & Public tests & $13.62$ & $100.0$ & $64.5$ & $87.9$ & $16.12$ & $100.0$ & $66.8$ & $82.5$ \\
LiveCodeBench & Public tests & $34.44$ & $100.0$ & $24.0$ & $89.2$ & $34.86$ & $100.0$ & $25.9$ & $88.0$ \\
CodeContests & Public tests & $41.18$ & $100.0$ & $20.7$ & $91.3$ & $43.10$ & $100.0$ & $19.7$ & $91.4$ \\
ZebraLogic & Constraint checks & $71.25$ & $100.0$ & $73.0$ & $34.8$ & $56.25$ & $100.0$ & $77.5$ & $56.2$ \\
OlympiadBench & LLM self-verifier & $73.78$ & $95.8$ & $46.3$ & $12.5$ & $74.37$ & $97.4$ & $44.4$ & $8.8$ \\
\bottomrule
\end{tabular}}
\vspace{-1.0em}
\end{table}

\subsection{Verifier Model and Examples}

In the unsupervised RLVR setting~\citep{ttrl2025,unsupervised_rlvr_2026}, the true correctness of sampled solutions is unobserved during training, so rewards must be constructed.
Prior work derives rewards from model-internal or sample-level statistics, including agreement among sampled outputs~\citep{ttrl2025} and verifier-free uncertainty objectives such as entropy minimization~\citep{emrl2025}.
We study a different yet common supervision source: manually specified error-detecting checks.
Such checks are often easy to define when full correctness is hard to verify, because they test necessary conditions rather than correctness itself.
Failing a check provides evidence that a solution is wrong, whereas passing the available checks only indicates that no checked failure mode was detected.

Let $x$ denote a problem instance and $y$ a complete candidate solution.
We distinguish between the latent ground-truth correctness
\[
r^*(x,y) \in \{0,1\},
\]
and the observed reward
\[
r(x,y) = V(x,y) \in \{0,1\},
\]
where $V$ denotes a manually specified error-detection scheme available at training time.
Such schemes are easy to obtain in many settings.
We describe three common instances below.

\paragraph{Executable checks.}
In coding tasks, public unit tests are commonly used to check basic correctness properties of a solution.
Given a set of input-output tests $\mathcal{T}_{\mathrm{pub}}(x)=\{(I_j,O_j)\}_{j=1}^{m}$, these tests define a binary reward function:
\begin{equation}
\label{eq:public-test-verifier}
V_{\mathrm{pub}}(x,y)
=
\mathbb{I}\!\left[
\forall (I_j,O_j)\in\mathcal{T}_{\mathrm{pub}}(x),\;
\mathrm{Exec}(y,I_j)=O_j
\right].
\end{equation}
Passing all public unit tests is only a necessary condition for correctness, not a sufficient one.

\paragraph{Constraint checks.}
For many realistic tasks, a correct solution must satisfy explicit constraints: structured outputs must follow the required format, assignments must respect domain restrictions, and intermediate claims must obey local consistency requirements.
Let $\mathcal{C}(x)=\{c_j(x,\cdot)\}_{j=1}^{m_c}$ denote such necessary constraints induced by the task specification.
These constraints define
\begin{equation}
\label{eq:constraint-verifier}
V_{\mathrm{con}}(x,y)
=
\mathbb{I}\!\left[
\forall c_j\in\mathcal{C}(x),\;
c_j(x,y)=1
\right].
\end{equation}
Failing a checked constraint is reliable evidence that the candidate is invalid, but satisfying all checked constraints does not certify that the full solution is correct.

\paragraph{LLM-as-verifier.}
For tasks with hard-to-formalize constraints, we use an LLM to search candidate solutions for errors rather than judge correctness, since certifying correctness is stronger than identifying a specific error.
Let $J_\phi(x,y)\in{\mathtt{Find},\mathtt{NotFind}}$ be its final binary judgment, where $\mathtt{Find}$ denotes an identified failure mode and $\mathtt{NotFind}$ denotes that none was found.
We define the corresponding reward as
\begin{equation}
\label{eq:llm-verifier}
V_{\mathrm{llm}}(x,y)
=
\mathbb{I}\!\left[J_\phi(x,y)=\mathtt{NotFind}\right].
\end{equation}
Unlike executable or deterministic constraint checks, an LLM-based signal provides no structural protection against false negatives, as it may reject valid solutions.
Thus, its one-sidedness must be empirically measured rather than assumed; Section~\ref{sec:measuring-one-sidedness} reports these measurements.

These examples instantiate the same broader problem.
The observed reward $r(x,y)$ is an imperfect proxy for $r^*(x,y)$, but its errors are asymmetric: negative outcomes are much more reliable than positive outcomes.
The central question in this paper is not how to eliminate verifier noise completely, but how to perform RL when the available signal can reliably falsify some bad solutions but cannot reliably certify good ones.

\subsection{Measuring One-Sidedness}
\label{sec:measuring-one-sidedness}

We measure this asymmetry with two conditional quantities:
\begin{align}
    q_{\mathrm{neg}}
    &\coloneqq
    \Pr(r^*=0 \mid r=0),
    \label{eq:qneg}
    \\
    q_{\mathrm{pos}}
    &\coloneqq
    \Pr(r^*=0 \mid r=1).
    \label{eq:qpos}
\end{align}
These quantities measure the probability that the ground-truth reward is false under each observed reward outcome: $q_{\mathrm{neg}}$ is the probability that a verifier-negative sample is truly incorrect, and $q_{\mathrm{pos}}$ is the probability that a verifier-positive sample is nevertheless incorrect.
We call a verifier \emph{one-sided verifiable} when $q_{\mathrm{neg}}$ is high but $q_{\mathrm{pos}}$ remains non-negligible.
In this regime, observed negatives can be used as trusted counterexamples, whereas observed positives should be treated as non-falsified unlabeled candidates rather than ground-truth positives.

Table~\ref{tab:osv} reports $q_{\mathrm{neg}}$, $q_{\mathrm{pos}}$, and negative-class recall
$r_{\mathrm{neg}} \coloneqq \Pr(r=0 \mid r^*=0)$, the fraction of truly incorrect responses flagged by the verifier.
For public tests and constraint checks, $q_{\mathrm{neg}}=1$ by construction because they check necessary conditions for correctness.
For the LLM self-verifier, $q_{\mathrm{neg}}$ is close to $1$, showing accurate error identification with few rejected correct answers.
Across all three verifier types, $q_{\mathrm{pos}}$ remains non-negligible, showing a substantial false-positive fraction among verifier-positive responses.
Yet $r_{\mathrm{neg}}$ varies by an order of magnitude: code public tests flag $82$--$91\%$ of true failures, ZebraLogic constraint checks $35$--$56\%$, and the LLM self-verifier on OlympiadBench only $9$--$13\%$.
Thus, OSV asymmetry persists under different per-group negative densities.
These quantities are computed using ground-truth labels $r^*$ solely for analysis and unavailable during training.\looseness=-1

\paragraph{Learning implications.}
This noise profile changes which verifier outcomes RLVR should trust.
Standard RLVR treats $r=1$ as a positive label and $r=0$ as a negative label.
Under OSV, however, the $r=1$ pool mixes true solutions with unchecked failures.
Reinforcing this pool gives shortcut trajectories a direct positive gradient whenever they survive the partial checks.

%% file: content/nopo.tex
\section{Negative-Only Policy Optimization}
\label{sec:method}

\begin{figure*}[t]
  \centering
  \includegraphics[width=\textwidth]{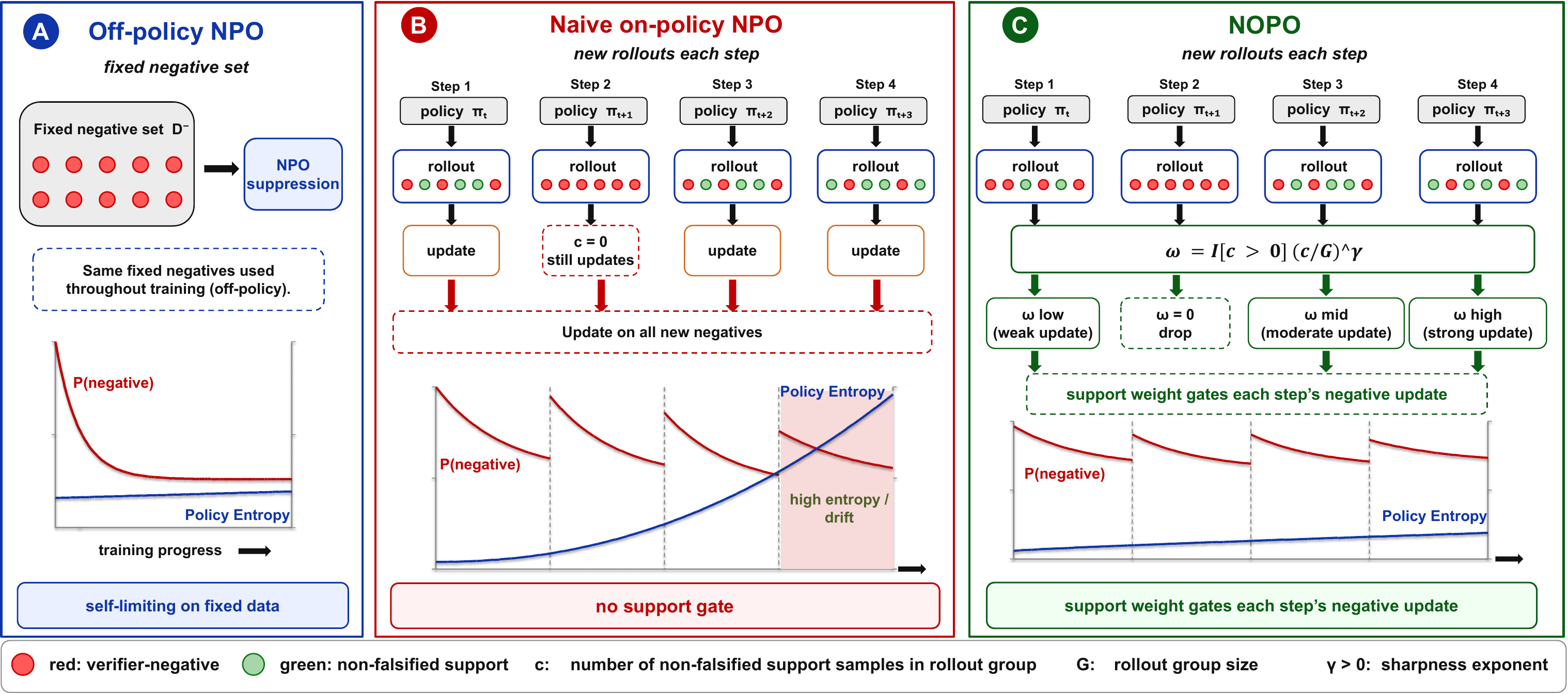}
  \caption{Overview of Negative-Only Policy Optimization (NOPO).}
  \label{fig:nopo}
  \vspace{-1.5em}
\end{figure*}

In the one-sided verifiable rewards (OSV) setting, the reliability of verifier-positive samples is generally unknown: $q_{\mathrm{pos}}$ is difficult to observe in realistic training environments.
We therefore learn only from the reliable side of the signal, namely verifier-negative samples.
This principle resembles Negative Preference Optimization (NPO), introduced in the preliminaries.
NPO-style unlearning methods aim to reduce the likelihood of specified undesirable examples or behaviors relative to a reference policy, while preserving the model's original capabilities.

However, NPO-style algorithms are typically designed for off-policy updates on a fixed negative dataset, unlike on-policy RLVR.
With fixed negatives, NPO suppresses undesirable examples relative to a reference policy and stops once their likelihood is sufficiently reduced.
In RLVR, however, negatives are generated online by the changing policy itself.
Negative-only updates therefore face a group-level ambiguity: suppressing rejected trajectories redistributes probability mass, but verifier-negative samples alone do not specify where it should move.
Overly aggressive suppression can induce unconstrained ``move-away'' updates, causing the policy to generate new failures that are then further suppressed.

Directly applying NPO is therefore insufficient for OSV-based RL training.
Although the NPO loss provides sample-level self-attenuation relative to the reference policy, it does not distinguish whether a verifier-negative rollout comes from a group that also contains non-falsified responses.
This distinction is important in on-policy RLVR: when the same rollout group contains non-falsified responses, suppressing clear failures is more likely to redistribute probability mass toward the non-falsified part of the sampled group.
To adapt NPO to this setting, we propose \emph{Negative-Only Policy Optimization} (NOPO), which modulates NPO suppression by the amount of non-falsified support in the same rollout group.
Importantly, verifier-positive samples are not treated as correct demonstrations.
NOPO uses only their aggregate count as a support signal, indicating whether the current policy has produced non-falsified responses under the available verifier.

\paragraph{Non-falsified support weight.}
Given a prompt $x$, let
$\mathbf{y}=(y_1,\ldots,y_G)$ be a rollout group sampled from the old policy
$\pi_{\theta_{\mathrm{old}}}(\cdot \mid x)$, and let
$r_i = V(x,y_i) \in \{0,1\}$ denote the verifier label of the $i$-th rollout.
We define the amount of non-falsified support in the group as
\begin{equation}
\label{eq:positive-count}
c(\mathbf{r}) \coloneqq \sum_{i=1}^{G} r_i ,
\end{equation}
where $\mathbf{r}=(r_1,\ldots,r_G)$.
If $c(\mathbf{r})=0$, the rollout group contains no non-falsified response under the available verifier, and negative-only suppression provides little information about a useful direction for probability redistribution.
This event is common on hard prompts: under the rollout distribution, $\Pr[c(\mathbf{r})=0]$ equals one minus the verifier-$\mathrm{pass}@G$ rate, namely the probability that none of the $G$ sampled responses passes the available verifier checks.

NOPO therefore weights each negative update by the non-falsified support within its rollout group:
\begin{equation}
\label{eq:group-weight}
\omega_{\gamma}(\mathbf{r})
\coloneqq
\begin{cases}
1, & \gamma = 0, \\[3pt]
\left(\dfrac{c(\mathbf{r})}{G}\right)^{\gamma}, & \gamma > 0,
\end{cases}
\qquad \gamma \ge 0 .
\end{equation}
The default $\gamma=1$ gives a linear support weight.
When $\gamma>0$, groups with no verifier-positive responses receive zero weight automatically, low-support groups contribute less, and groups with more non-falsified support allow stronger suppression of verifier-negative responses.
Setting $\gamma=0$ recovers vanilla NPO, where all groups contribute uniformly, while larger $\gamma$ makes NOPO more conservative on low-support groups.

The resulting NOPO objective applies NPO-style suppression only to verifier-negative responses, modulated by the group-level non-falsified support weight:
\begin{equation}
\label{eq:nopo-objective}
\mathcal{L}_{\mathrm{NOPO}}(\theta)
=
\mathbb{E}_{
x\sim\mathcal{D},
\mathbf{y}\sim\pi_{\theta_{\mathrm{old}}}(\cdot\mid x)
}
\left[
\omega_{\gamma}(\mathbf{r})
\cdot
\frac{1}{G}
\sum_{i=1}^{G}
\mathbb{I}[r_i=0]\,
\mathcal{L}_{\mathrm{NPO}}(\theta;x,y_i)
\right],
\end{equation}
where $r_i=V(x,y_i)$ and $\mathcal{L}_{\mathrm{NPO}}$ is the reference-relative NPO loss with temperature $\beta$.
The verifier labels and the support weight are treated as fixed feedback for the sampled rollout group; gradients are applied only through the NPO loss on verifier-negative responses.
Algorithm~\ref{alg:nopo} in Appendix~\ref{app:nopo-algorithm} summarizes the corresponding training update.

This formulation separates reliability at two levels.
The NPO term provides sample-level control: its gradient self-attenuates once the current policy assigns a negative response lower probability than the reference.
The group weight $\omega_{\gamma}(\mathbf{r})$ provides group-level control: it trusts negative evidence only when the same on-policy rollout group contains verifier-positive support.
As a result, NOPO updates on negative rollouts only when the group indicates that the prompt is within the model's current reach, avoiding unconstrained ``move-away'' updates on prompts where all sampled trajectories fail.

%% file: content/experiments.tex
\section{Experiments}
\label{sec:experiment}

\subsection{Setup}

We use Qwen3-4B and Qwen3-8B as base models~\citep{qwen3}.
Code experiments cover LiveCodeBench~\citep{livecodebench}, TACO test~\citep{taco}, and CodeContests~\citep{alphacode} with public-test binary rewards, with CodeForces reserved as an unseen cross-benchmark transfer target.
Logic experiments use ZebraLogic with a deterministic constraint checker that verifies format and domain constraints without certifying full correctness, while math experiments use OlympiadBench~\citep{olympiadbench} with an LLM self-verifier binary reward.
We compare four methods, each using $G=16$ samples per prompt: \textbf{GRPO}, which directly optimizes the observed verifier reward; \textbf{EM-RL}, a token-level verifier-free baseline that minimizes policy entropy during training~\citep{emrl2025}; \textbf{w-Reinforce}, which assigns verifier-negative samples greater weight than positive ones~\citep{nsr}; and \textbf{NOPO}, our proposed method.

\subsection{Main Result}
\label{sec:main-result}

\begin{table}[t]
\centering
\small
\caption{Main results across code, logic, and math benchmarks under one-sided verifiable rewards.}
\label{tab:main-result}
\setlength{\tabcolsep}{4.5pt}
\renewcommand{\arraystretch}{1.05}
\begin{tabular}{lcccccc}
\toprule
\textbf{Method} & \textbf{TACO} & \textbf{CodeContests} & \textbf{LCB} & \textbf{ZebraLogic} & \textbf{Olympiad} & \textbf{Avg.} \\
\midrule
\multicolumn{7}{l}{\textit{Qwen3-4B}} \\
Base           & 13.62 & 41.18 & 34.44 & 71.25 & 73.78 & 46.85 \\
GRPO           & 15.89 & 43.75 & 36.17 & 74.38 & 74.44 & 48.93 \\
EM-RL          & 15.44 & 40.91 & 35.26 & 69.37 & \underline{74.89} & 47.17 \\
w-Reinforce    & \underline{18.62} & \underline{47.63} & \underline{38.52} & \underline{75.00} & 73.70 & \underline{50.69} \\
\rowcolor{gray!15} \textbf{NOPO} & \textbf{19.52} & \textbf{50.47} & \textbf{39.61} & \textbf{79.37} & \textbf{75.33} & \textbf{52.86} \\
\midrule
\multicolumn{7}{l}{\textit{Qwen3-8B}} \\
Base           & 16.12 & 43.10 & 34.86 & 56.25 & 74.37 & 44.94 \\
GRPO           & 17.03 & 49.58 & 37.14 & \underline{60.62} & \textbf{76.97} & 48.27 \\
EM-RL          & 16.23 & 44.96 & 34.86 & 53.75 & 74.74 & 44.91 \\
w-Reinforce    & \underline{18.50} & \underline{50.42} & \underline{41.14} & 59.38 & 74.74 & \underline{48.84} \\
\rowcolor{gray!15} \textbf{NOPO} & \textbf{19.98} & \textbf{55.04} & \textbf{42.86} & \textbf{66.25} & \underline{76.52} & \textbf{52.13} \\
\bottomrule
\end{tabular}
\vspace{-1.2em}
\end{table}

As shown in Table~\ref{tab:main-result}, NOPO achieves the best average $\text{pass}@16$ on both backbones among RL methods, with w-Reinforce second on average.
Both methods treat verifier-negative samples as trustworthy supervision but differ in how they control policy drift from negative-only penalties: w-Reinforce gives verifier-positive samples a small positive weight to offset the dominant negative gradient, whereas NOPO directly rescales each prompt's negative-only loss by its within-group non-falsified support.
By contrast, verifier-free EM-RL, which uses only token-level entropy minimization, gives essentially no gain over Base on either backbone, consistent with the view that intrinsic-reward methods are bounded by sharpening the base policy's initial distribution~\citep{emrl2025,rlif2025,yue2025limit,unsupervised_rlvr_2026}.
At the per-cell level, NOPO ranks first in $9$ of $10$ benchmark--backbone cells; the only exception is OlympiadBench under Qwen3-8B, where it ranks second behind GRPO.
As Table~\ref{tab:osv} shows, OlympiadBench has the lowest verifier negative-class recall in our setup, so few truly incorrect rollouts are flagged, weakening NOPO's negative-only supervision signal.\looseness=-1

\subsection{Inference-Time Scaling}
\label{sec:scaling}

A central debate around RLVR is whether it expands a model's capability frontier~\citep{prorl2025} or merely concentrates probability mass on solutions already discoverable at large $k$~\citep{yue2025limit,evolrl2025}.
We therefore test whether our gains persist at large inference budgets, rather than reflecting sharpening around solutions reachable by the base model with enough samples.
Table~\ref{tab:scaling-pass-at-k-4b} reports Qwen3-4B on CodeContests and LiveCodeBench as $k$ scales from $1$ to $512$.
NOPO shows an advantage at larger inference budgets, and this advantage \emph{widens} with $k$, indicating genuine frontier expansion: the post-RL policy solves problems the base model cannot reach even with hundreds of attempts.
w-Reinforce, which also suppresses verifier-negative outputs, is the consistent runner-up at large $k$, supporting the OSV view that trusted-negative suppression drives frontier expansion.
NOPO's per-prompt non-falsified support weight further improves on this principle.
GRPO, by contrast, illustrates the sharpening failure mode: its $\text{pass}@1$ is the highest across methods, but at $k=512$ it falls \emph{below} the base model on both benchmarks, indicating that small-$k$ gains come at the cost of exploration.
More encouragingly, NOPO's frontier-expanding behavior suggests that modulating negative-only updates by per-prompt non-falsified support may transfer to other RLVR settings as a remedy for entropy collapse.\looseness=-1

\begin{table}[t]
\centering
\small
\caption{Inference-time $\text{pass}@k$ on Qwen3-4B. \textbf{Bold} marks the maximum, \underline{underline} the second-best.}
\label{tab:scaling-pass-at-k-4b}
\resizebox{\linewidth}{!}{
\begin{tabular}{lccccccccc}
\toprule
\textbf{Method} & $k=1$ & $k=4$ & $k=8$ & $k=16$ & $k=32$ & $k=64$ & $k=128$ & $k=256$ & $k=512$ \\
\midrule
\multicolumn{10}{l}{\textit{CodeContests}} \\
Base           & 25.04 & 33.51 & 37.50 & 41.18 & 44.55 & 47.35 & 49.21 & 50.29 & 51.05 \\
GRPO           & \textbf{37.23} & \textbf{40.74} & 42.35 & 43.75 & 44.92 & 45.92 & 46.82 & 47.72 & 48.54 \\
EM-RL          & 25.08 & 33.36 & 37.30 & 40.91 & 44.23 & 47.09 & 49.15 & 50.34 & 51.46 \\
w-Reinforce    & 22.39 & 37.01 & \underline{42.80} & \underline{47.63} & \underline{52.08} & \underline{56.16} & \underline{59.61} & \underline{62.08} & \underline{63.60} \\
\rowcolor{gray!15} \textbf{NOPO (ours)} & \underline{26.87} & \underline{40.58} & \textbf{45.84} & \textbf{50.47} & \textbf{54.48} & \textbf{57.89} & \textbf{61.03} & \textbf{64.11} & \textbf{66.95} \\
\midrule
\multicolumn{10}{l}{\textit{LiveCodeBench}} \\
Base           & 26.03 & 31.11 & 32.89 & 34.44 & 35.93 & 37.30 & 38.58 & 40.07 & 42.29 \\
GRPO           & \textbf{29.40} & 33.07 & 34.58 & 36.17 & 37.66 & 38.96 & 40.13 & 41.03 & 41.71 \\
EM-RL          & 25.74 & 31.41 & 33.39 & 35.26 & 37.04 & 38.61 & 40.05 & 41.56 & 43.43 \\
w-Reinforce    & \underline{26.21} & \underline{33.45} & \underline{36.21} & \underline{38.52} & \underline{40.55} & \underline{42.64} & \underline{44.74} & \underline{46.75} & \underline{48.57} \\
\rowcolor{gray!15} \textbf{NOPO (ours)} & 25.33 & \textbf{33.72} & \textbf{36.75} & \textbf{39.61} & \textbf{42.34} & \textbf{44.89} & \textbf{47.04} & \textbf{48.88} & \textbf{50.86} \\
\bottomrule
\end{tabular}}
\vspace{-1.2em}
\end{table}

\subsection{Policy generalization}
\label{sec:generalization}

To test whether OSV-trained gains transfer beyond the source benchmark, Table~\ref{tab:transfer-pass-at-16-4b} evaluates checkpoints trained on CodeContests and LiveCodeBench on unseen code benchmarks.
GRPO again shows limited generalization: its main-result gains fail to transfer even within the same domain, suggesting source-benchmark overfitting rather than genuine capability improvement.
By contrast, w-Reinforce and NOPO, both focused on verifier-negative samples, retain clear transfer gains, and NOPO is best on every unseen target.
This highlights the advantage of effectively leveraging negative samples under OSV.
EM-RL, lacking a sustained effective optimization signal, performs on par with Base.

\begin{table}[t]
\centering
\small
\caption{Cross-bench transfer $\text{pass}@16$ on Qwen3-4B with two training sources. \textbf{Bold} marks the row maximum and \underline{underline} the second-best among RL methods.}
\label{tab:transfer-pass-at-16-4b}
\setlength{\tabcolsep}{6pt}
\renewcommand{\arraystretch}{1.05}
\begin{tabular}{l l c c c c >{\columncolor{gray!15}}c}
\toprule
\textbf{Source} & \textbf{Target} & \textbf{Base} & \textbf{GRPO} & \textbf{EM-RL} & \textbf{w-Reinforce} & \textbf{NOPO (ours)} \\
\midrule
\multirow{4}{*}{CodeContests}
 & TACO          & 13.62 & 12.71 & 13.28 & \underline{14.42} & \textbf{15.78} \\
 & LiveCodeBench & 34.44 & 32.57 & 34.29 & \underline{38.86} & \textbf{40.57} \\
 & CodeForces    & 22.91 & 24.63 & 23.13 & \underline{29.76} & \textbf{32.12} \\
\cmidrule(l){2-7}
 & \textit{Avg-Unseen} & 23.66 & 23.30 & 23.57 & \underline{27.68} & \textbf{29.49} \\
\midrule
\multirow{4}{*}{LiveCodeBench}
 & TACO          & 13.62 & 12.49 & 13.62 & \underline{15.55} & \textbf{16.00} \\
 & CodeContests  & 41.18 & 39.75 & 43.10 & \underline{47.70} & \textbf{49.37} \\
 & CodeForces    & 22.91 & 22.48 & 24.84 & \underline{29.34} & \textbf{30.41} \\
\cmidrule(l){2-7}
 & \textit{Avg-Unseen} & 25.90 & 24.91 & 27.19 & \underline{30.86} & \textbf{31.93} \\
\bottomrule
\end{tabular}
\vspace{-1.em}
\end{table}

\subsection{Ablation: support exponent \texorpdfstring{$\gamma$}{gamma} and naive baselines}
\label{sec:ablation}

\begin{wrapfigure}{r}{0.5\textwidth}
   \vspace{-1.2em}
   \centering
   \includegraphics[width=0.48\textwidth]{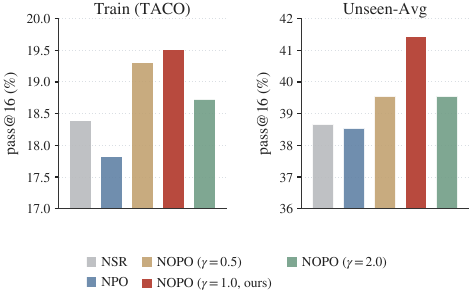}
   \caption{Ablation study on Qwen3-4B trained on TACO and evaluated on unseen benchmarks}
   \label{fig:ablation-4b-taco}
   \vspace{-1.2em}
\end{wrapfigure}

This ablation tests the necessity of the non-falsified support weight and its robustness to the support exponent $\gamma$.
Here, \emph{NPO} directly applies the NPO loss in RL, while \emph{NSR} is w-Reinforce without verifier-positive optimization.
We compare $\gamma \in \{0.5, 1.0, 2.0\}$, with $\gamma=1.0$ as the default.
All variants are trained on TACO with Qwen3-4B and evaluated on three unseen code targets.
Non-falsified support weights outperform both NSR and NPO in-domain and under cross-benchmark transfer, confirming their necessity.
The linear support weight ($\gamma=1.0$) performs best among the three exponents.
Detailed transfer results are in Table~\ref{tab:ablation-4b-taco-detail}.

%% file: content/related.tex
\section{Related Work}
\label{sec:related}

\paragraph{Unsupervised RLVR.}
Unsupervised RLVR can be categorized into two regimes by reward source~\citep{unsupervised_rlvr_2026}.
The \emph{intrinsic} regime derives rewards from the model itself through certainty- or consistency-based signals~\citep{ttrl2025,t3rl2026,emrl2025,evolrl2025,rlif2025,rzero2025}, and is provably bounded by the sharpening of the policy's initial distribution.
The \emph{external} regime grounds rewards in mechanisms outside the model and typically relies on verifiers that are effectively sound and complete, such as formal proof checkers, deterministic execution against complete test suites, or exact symbolic checks~\citep{hubert2025olympiad,absolutezero2025,ladder2025}.
Our setting falls within this external regime but relaxes its soundness-and-completeness assumption: readily available external verifiers in realistic deployment settings often cannot certify correctness, but can reliably reject visible failures, an asymmetry we formalize as one-sided verifiability.

\paragraph{RLVR with Noisy Rewards.}
A growing line of work studies RLVR when the verifier is imperfect~\citep{rateorfate2026,cai2025imperfect,elmansouri2025grpo}, and existing methods share a common recipe: characterize the verifier's noise---either through an aggregate quality index or through separately modeled false-positive and false-negative rates---and plug this characterization into the policy gradient as a correction or weighting term.
Obtaining such a characterization itself requires ground-truth labels for calibration, which our setting explicitly disallows; moreover, the rate at which the verifier falsely accepts incorrect outputs drifts upward as the policy learns to satisfy partial checks, so calibrated pre-training estimates or online noise-rate estimates may be unreliable in the OSV regime.
NOPO instead works structurally rather than statistically: it neither estimates nor corrects for noise rates, using the verifier only for the high-confidence negatives it can safely provide, consistent with the broader observation that false-positive rewards are particularly harmful as they reinforce shortcut behaviors~\citep{darkrich2024}.

\paragraph{Negative-Oriented Optimization.}
NOPO is also related to negative-oriented optimization.
LLM unlearning removes undesirable knowledge or behaviors while preserving general utility~\citep{eldan2023whos,maini2024tofu,li2024wmdp}.
Negative preference objectives such as NPO suppress a fixed off-policy forget set while avoiding naive gradient-ascent instability, with later variants revisiting the reference model~\citep{npo2024,fan2024simplicity}.
NOPO adopts NPO's self-attenuating suppression principle but applies it to on-policy verifier-negative rollouts, which introduces a redistribution ambiguity: negatives specify what to avoid, not where probability mass should move.
This motivates our group-level non-falsified support gate.
Another related line, negative sample reinforcement in RLVR, shows that penalizing incorrect rollouts can improve reasoning and preserve diversity~\citep{nsr,tang2025rethinking,ngrpo2025,lens2025}.
Unlike these methods, NOPO targets OSV, where verifier positives are unlabeled non-falsified support rather than reliable positive rewards.

%% file: content/conclusion.tex
\section{Conclusion}
\label{sec:conclusion}

As LLM capabilities improve and their applications expand, constructing complete verifiers for increasingly diverse training data becomes increasingly costly.
This paper studies a more practical setting where only partial verification is available.
We show that verifier rewards in this setting often exhibit one-sided verifiability: rejected samples provide reliable evidence of failure, whereas accepted samples are merely non-falsified rather than certified correct.
Under this asymmetry, directly treating verifier-positive samples as positive rewards introduces substantial noise and may encourage reward hacking rather than expand the model's true capability frontier.
We therefore propose Negative-Only Policy Optimization (NOPO), which learns only from high-confidence negative evidence.
NOPO combines sample-level reference-relative NPO suppression with group-level non-falsified support weighting, enabling controlled negative updates without directly reinforcing noisy positives.
Empirically, NOPO improves performance across multiple verifier types, with advantages that persist under inference-time scaling and cross-benchmark transfer.
These results suggest that properly leveraging negative evidence can substantially expand the model's capability frontier without ground-truth verifiers, making negative-only optimization a promising direction for scalable RLVR.

\paragraph{Limitations.}
Negative-only learning is inherently delicate: it specifies what the policy should avoid but not where removed probability mass should move.
This limitation motivates NOPO's conservative design and suggests future extensions.
In different one-sided verifiability scenarios, verifier-positive samples may contain exploitable structure, such as pairwise comparisons or uncertainty-based filtering, that could yield partial positive signals.
Incorporating such signals while preserving the robustness of negative-only optimization remains an important direction for future work.

%% file: appendix.tex
\newpage

\section{Broader impacts.}
\label{boarder_impacts}
A practical benefit of OSV and NOPO is that they can reduce the need for expensive complete verifiers, making RLVR-style training more accessible in domains where only partial checks are available.
At the same time, stronger reasoning and code-generation checkpoints can be misused, and optimization against incomplete checks can amplify unchecked failure modes if deployed without careful evaluation.

\section{NOPO training update}
\label{app:nopo-algorithm}

Algorithm~\ref{alg:nopo} gives the training update used for Negative-Only Policy Optimization.

\begin{algorithm}[h]
\caption{NOPO training update.}
\label{alg:nopo}
\begin{algorithmic}[1]
\State \textbf{Input:} dataset $\mathcal{D}$, policy $\pi_\theta$, reference policy $\pi_{\mathrm{ref}}$, verifier $V$, group size $G$, support exponent $\gamma$, NPO temperature $\beta$
\For{each training step}
    \State Sample prompt $x \sim \mathcal{D}$
    \State Set $\pi_{\theta_{\mathrm{old}}} \leftarrow \pi_\theta$ and generate rollouts $\mathbf{y}=(y_1,\ldots,y_G)$ with $y_i \sim \pi_{\theta_{\mathrm{old}}}(\cdot\mid x)$
    \State Compute verifier labels $r_i = V(x,y_i)$ for $i=1,\ldots,G$
    \State Compute $c(\mathbf{r})=\sum_{i=1}^{G} r_i$ and $\omega_{\gamma}(\mathbf{r})$ by Eq.~\ref{eq:group-weight}
    \State Take a gradient step on $\mathcal{L}_{\mathrm{NOPO}}(\theta)$ in Eq.~\ref{eq:nopo-objective}
\EndFor
\State \textbf{Output:} trained policy $\pi_\theta$
\end{algorithmic}
\end{algorithm}

\section[Experimental Details]{Experimental Details}
\label{app:experimental-details}

\subsection{Benchmark}
\label{app:benchmark-sizes}

All datasets involved are publicly available.
For LiveCodeBench, we use the code-generation split of \texttt{release\_v6}.
For TACO, we use the evaluation set released by TCS~\citep{xu2026tcs} and exclude 64 problems for which our parser cannot extract public test cases.
For CodeContests, we use the evaluation set released by CURE~\citep{cure2025}.
For ZebraLogic, we use the Medium-Large subset ($4\times4$, $4\times5$, $5\times3$, and $6\times2$).
Unless otherwise specified above, datasets are used as provided.
For the main results in Table~\ref{tab:main-result}, training and evaluation use the same test set for each benchmark, whereas the transfer experiments in Table~\ref{tab:transfer-pass-at-16-4b} evaluate on target benchmarks not used for training.
Only one-sided verifier rewards are available during training, and all methods are evaluated using the oracle.
Table~\ref{tab:train-oracle} summarizes the training verifiers and evaluation oracles.

\begin{table}[htbp]
\centering
\small
\caption{Training rewards, evaluation oracles, and mean test counts per problem.}
\label{tab:train-oracle}
\setlength{\tabcolsep}{4pt}
\begin{tabular}{p{0.15\linewidth}p{0.27\linewidth}p{0.28\linewidth}p{0.18\linewidth}}
\toprule
Benchmark & Training reward & Evaluation oracle & Tests: OSV / oracle \\
\midrule
TACO & public-test binary reward & full test suite (public + hidden) & 1.7 / 204 \\
CodeContests & public-test binary reward & full test suite (public + hidden) & 1.9 / 12 \\
LCB & public-test binary reward & full test suite (public + hidden) & 2.6 / 40 \\
ZebraLogic & constraint checker & ground-truth answer & -- \\
OlympiadBench & LLM self-verifier & ground-truth answer & -- \\
\bottomrule
\end{tabular}
\end{table}

\subsection{Experimental Configuration}
\label{app:experimental-configuration}

All RL runs use the verl framework~\citep{hybridflow} with vLLM rollouts~\citep{vllm} and an FSDP actor.
The two backbones are Qwen3-4B and Qwen3-8B in non-thinking mode.
Unless noted otherwise, every method shares the optimisation, sampling, and bookkeeping settings below; only the loss term and its method-specific hyperparameter change between methods.

\paragraph{Optimisation.}
We use AdamW with a constant learning rate of $5\!\times\!10^{-7}$ after a $20$-step linear warmup.
The training batch is $16$ prompts; each prompt produces $G=16$ on-policy rollouts, giving a $16{\times}16=256$-rollout step.
PPO mini-batches contain $4$ prompts and we run a single inner PPO epoch per global step. The maximum number of global steps is $150$.
For all methods dominated by negative optimisation, prolonged training may lead to entropy divergence, so we monitor policy entropy during training and retain the most recent checkpoint saved before an observed entropy divergence for evaluation.

\paragraph{Sampling.}
A single sampling profile is shared between RL rollouts and validation rollouts.
For Qwen3 models in non-thinking mode, generation uses \texttt{temperature}{=}$0.8$, \texttt{top\_p}{=}$0.8$, \texttt{top\_k}{=}$20$, and \texttt{min\_p}{=}$0$.

\paragraph{Sequence lengths.}
Code experiments use \texttt{max\_response\_length}{=}$8192$.
Logic and math experiments use \texttt{max\_response\_length}{=}$16384$.

\paragraph{Method-specific hyperparameters.}
\begin{itemize}
    \item \textbf{GRPO}: $\text{kl\_loss\_coef}{=}10^{-3}$, no other deviations from the verl default.
    \item \textbf{w-Reinforce}~\citep{nsr}: verifier-positive samples receive weight $\lambda=0.1$, verifier-negative samples receive weight $1$.
    \item \textbf{EM-RL}~\citep{emrl2025}: \texttt{adv\_estimator}{=}\texttt{rloo}; the training reward is replaced by token-level entropy minimisation computed inside the trainer from actor log-probabilities, and the environment reward manager returns zero placeholders on the training path. Validation still routes through the task-native evaluator so $\text{pass}@k$ is reported on the same scale as the other methods.
    \item \textbf{NPO}: $\beta=0.1$, \texttt{kl\_loss\_coef}{=}$0$. The reference policy $\pi_{\mathrm{ref}}$ is the frozen base model.
    \item \textbf{NOPO} (default): $\beta=0.1$, $\gamma=1.0$, \texttt{kl\_loss\_coef}{=}$0$. The non-falsified support weight $\omega_{\gamma}(\mathbf{r})$ from Eq.~\ref{eq:group-weight} is computed from the same on-policy rollout group used for the NPO term; $c(\mathbf{r}){=}0$ groups receive zero weight automatically and are dropped from the actor update.
\end{itemize}

\paragraph{Inference-time scaling.}
For the inference-scaling sweep (Table~\ref{tab:scaling-pass-at-k-4b}), we resample $n=512$ rollouts per problem from each method's checkpoint with the same sampling profile and report $\text{pass}@k$ as a function of $k\in\{1,4,8,16,32,64,128,256,512\}$.

\paragraph{Compute resources.}
All RL training runs were executed on eight NVIDIA A100 GPUs with 80GB memory per GPU.
A typical (method, backbone, benchmark) run takes approximately 12 hours under this setup.
Evaluation-only sweeps reuse selected checkpoints and are cheaper than training, although the inference-time scaling results in Table~\ref{tab:scaling-pass-at-k-4b} require generating up to 512 rollouts per problem and still take approximately 12 hours under the same configuration.

\section{Per-target ablation numbers}
\label{app:ablation-4b-taco}

Table~\ref{tab:ablation-4b-taco-detail} expands Figure~\ref{fig:ablation-4b-taco} with per-target $\text{pass}@16$ for the five-method ablation on Qwen3-4B trained on TACO.
\textbf{Train} is in-domain $\text{pass}@16$; the remaining columns are unseen-transfer $\text{pass}@16$ ($G=16$ inference samples per problem) for the same selected checkpoint.
\textbf{Unseen-Avg} is the macro mean of the three unseen-transfer columns and matches the right-hand bars of Figure~\ref{fig:ablation-4b-taco}.
\textbf{Bold} marks the column maximum.

\begin{table}[h]
\centering
\small
\caption[Per-target $\text{pass}@16$ for the ablation on Qwen3-4B trained on TACO.]{Per-target $\text{pass}@16$ for the Section~\ref{sec:ablation} ablation on Qwen3-4B trained on TACO.}
\label{tab:ablation-4b-taco-detail}
\setlength{\tabcolsep}{5pt}
\renewcommand{\arraystretch}{1.1}
\begin{tabular}{l c c c c c}
\toprule
\textbf{Method} & \textbf{Train} & \textbf{LiveCodeBench} & \textbf{CodeContests} & \textbf{CodeForces} & \textbf{Unseen-Avg} \\
\midrule
NSR                          & 18.39 & 37.71 & 48.95 & 29.34 & 38.67 \\
NPO                          & 17.82 & 37.71 & 47.70 & 30.19 & 38.53 \\
NOPO ($\gamma = 0.5$)        & 19.30 & 38.86 & 48.54 & 31.26 & 39.55 \\
\rowcolor{gray!15} \textbf{NOPO ($\gamma = 1.0$, default)} & \textbf{19.52} & \textbf{39.43} & \textbf{51.05} & \textbf{33.83} & \textbf{41.44} \\
NOPO ($\gamma = 2.0$)        & 18.73 & 38.86 & 48.95 & 30.84 & 39.55 \\
\bottomrule
\end{tabular}
\end{table}

\section{Post-training verifier noise}
\label{app:postrl}
Table~\ref{tab:postrl-noise} reports post-RL verifier noise for Qwen3-4B trained on CodeContests, with Base statistics from Table~\ref{tab:osv} and unseen-target averages from Table~\ref{tab:transfer-pass-at-16-4b}.
Public tests check necessary conditions for correctness, so $q_{\mathrm{neg}}$ remains $100\%$ across all policies.
GRPO increases $q_{\mathrm{pos}}$ and decreases $r_{\mathrm{neg}}$, indicating more verifier-escaping errors.
In contrast, w-Reinforce and NOPO reduce $q_{\mathrm{pos}}$ while maintaining high $r_{\mathrm{neg}}$, with NOPO achieving the lowest $q_{\mathrm{pos}}$ and the highest in-domain and average transfer performance.

\begin{table}[htbp]
\centering
\small
\caption{Post-RL verifier noise and oracle performance on CodeContests with Qwen3-4B. All values are percentages.}
\label{tab:postrl-noise}
\setlength{\tabcolsep}{4pt}
\begin{tabular}{lccccc}
\toprule
Policy & $q_{\mathrm{neg}}$ & $q_{\mathrm{pos}}$ & $r_{\mathrm{neg}}$ & In-domain $\mathrm{pass}@16$ & Transfer average \\
\midrule
Base & 100.0 & 20.7 & 91.3 & 41.18 & 23.66 \\
GRPO & 100.0 & 25.2 & 79.8 & 43.75 & 23.30 \\
w-Reinforce & 100.0 & 18.1 & 93.4 & 47.63 & 27.68 \\
NOPO & 100.0 & 16.6 & 92.4 & 50.47 & 29.49 \\
\bottomrule
\end{tabular}
\end{table}

\section{Per-problem overlap at a fixed inference budget}
\label{app:overlap}
We define the capability frontier operationally as the set of problems solved under a fixed inference budget.
Following the Qwen3-4B scaling experiment in Table~\ref{tab:scaling-pass-at-k-4b}, a problem is counted as solved if at least one of its $512$ oracle-evaluated samples is correct.
Table~\ref{tab:overlap} shows that NOPO retains $120/122$ Base-solved CodeContests problems and $73/74$ Base-solved LiveCodeBench problems, while additionally solving $40$ and $16$ problems, respectively.
Thus, NOPO nearly preserves the Base-solved set while enlarging the reachable problem set under the same sampling budget.

\begin{table}[htbp]
\centering
\small
\caption{Problem-level solution overlap at $512$ samples per problem, Qwen3-4B.}
\label{tab:overlap}
\setlength{\tabcolsep}{4pt}
\begin{tabular}{lrrrrrrr}
\toprule
Benchmark & Total & Base solved & NOPO solved & Both & Neither & Base-only & NOPO-only \\
\midrule
CodeContests & 239 & 122 & 160 & 120 & 77 & 2 & 40 \\
LiveCodeBench & 175 & 74 & 89 & 73 & 85 & 1 & 16 \\
\bottomrule
\end{tabular}
\end{table}

\section{Verifier characteristics and NOPO gains}
\label{app:asymmetry-gains}
Table~\ref{tab:asymmetry-gains} compares verifier asymmetry with NOPO's $\mathrm{pass}@16$ gains over GRPO across all ten settings.
NOPO improves over GRPO at both extremes of $q_{\mathrm{pos}}$ ($19.7$--$77.5\%$): non-falsified support scales negative updates rather than directly reinforcing verifier-positive responses.
The clearest association is with negative recall: NOPO gains $2.95$--$6.72$ points in the eight code/logic settings, where verifiers detect at least a third of true failures.
On OlympiadBench, recall is only $9$--$13\%$, leaving few trusted negatives and reducing the NOPO--GRPO margins to $+0.89$ and $-0.45$ points.

\begin{table}[htbp]
\centering
\small
\caption{Pre-RL verifier statistics and the NOPO--GRPO $\mathrm{pass}@16$ gap. Parentheses give the relative improvement over GRPO.}
\label{tab:asymmetry-gains}
\setlength{\tabcolsep}{4pt}
\begin{tabular}{llccccc}
\toprule
Benchmark & Backbone & $q_{\mathrm{pos}}$ & $r_{\mathrm{neg}}$ & GRPO & NOPO & NOPO--GRPO \\
\midrule
CodeContests & 4B & 20.7 & 91.3 & 43.75 & 50.47 & +6.72 (+15.4\%) \\
CodeContests & 8B & 19.7 & 91.4 & 49.58 & 55.04 & +5.46 (+11.0\%) \\
LCB & 4B & 24.0 & 89.2 & 36.17 & 39.61 & +3.44 (+9.5\%) \\
LCB & 8B & 25.9 & 88.0 & 37.14 & 42.86 & +5.72 (+15.4\%) \\
TACO & 4B & 64.5 & 87.9 & 15.89 & 19.52 & +3.63 (+22.8\%) \\
TACO & 8B & 66.8 & 82.5 & 17.03 & 19.98 & +2.95 (+17.3\%) \\
ZebraLogic & 4B & 73.0 & 34.8 & 74.38 & 79.37 & +4.99 (+6.7\%) \\
ZebraLogic & 8B & 77.5 & 56.2 & 60.62 & 66.25 & +5.63 (+9.3\%) \\
OlympiadBench & 4B & 46.3 & 12.5 & 74.44 & 75.33 & +0.89 (+1.2\%) \\
OlympiadBench & 8B & 44.4 & 8.8 & 76.97 & 76.52 & $-$0.45 ($-$0.6\%) \\
\bottomrule
\end{tabular}
\end{table}

\section{Comparison with NPO}
\label{app:npo-comparison}
Table~\ref{tab:npo-five-bench} extends the NPO comparison to all five benchmarks on Qwen3-8B under the main experimental protocol.
NOPO improves over NPO, the $\gamma=0$ special case of Eq.~\ref{eq:group-weight}, on every benchmark by $1.19$--$4.37$ points, supporting the contribution of the non-falsified support weight.
NPO also outperforms GRPO on average ($49.35$ versus $48.27$).

\begin{table}[htbp]
\centering
\small
\caption{Direct NPO comparison on Qwen3-8B. GRPO and NOPO results are reproduced from Table~\ref{tab:main-result}.}
\label{tab:npo-five-bench}
\setlength{\tabcolsep}{4pt}
\begin{tabular}{lcccccc}
\toprule
Method & TACO & CodeContests & LCB & ZebraLogic & Olympiad & Avg. \\
\midrule
GRPO & 17.03 & 49.58 & 37.14 & 60.62 & 76.97 & 48.27 \\
NPO & 18.16 & 52.52 & 38.86 & 61.88 & 75.33 & 49.35 \\
NOPO & 19.98 & 55.04 & 42.86 & 66.25 & 76.52 & 52.13 \\
\bottomrule
\end{tabular}
\end{table}

For a rollout group with verifier-negative fraction $p$, the total suppression weight before the common $1/G$ normalization scales as $pG$ for NPO and $p(1-p)G$ for NOPO with $\gamma=1$.
NPO therefore applies the largest suppression to all-failed groups, whereas NOPO's weight is maximal for mixed groups and vanishes at both extremes.
For example, groups with $80\%$ and $20\%$ negative rollouts receive equal total weight under NOPO, while NPO assigns four times as much to the former.

\section{Verifier Implementation Details}
\label{app:verifier-details}

This section specifies the two non-trivial verifiers used in the main results: the ZebraLogic constraint-check verifier (Section~\ref{app:verifier-zebralogic}) and the LLM self-verifier on OlympiadBench (Section~\ref{app:verifier-llm-self}).

\subsection{Constraint-Check Verifier on ZebraLogic}
\label{app:verifier-zebralogic}

ZebraLogic puzzles ask the model to assign $K-1$ features to $N$ houses subject to a list of natural-language clues; the gold solution is a JSON object \texttt{\{"header": [...], "rows": [[...]]\}} with exactly $N$ rows.
Our verifier checks whether the candidate assignment satisfies the puzzle's \emph{schema and domain constraints} — necessary conditions induced by the task specification — without consulting the clues or the gold assignment.

\subsection{LLM Self-Verifier on OlympiadBench}
\label{app:verifier-llm-self}

For OlympiadBench, the policy and the verifier use the \emph{same} Qwen3-4B or Qwen3-8B backbone in non-thinking mode.

\paragraph{Three-turn audit prompt.}
For each rollout we send a three-turn chat:
\begin{enumerate}
    \item \texttt{user}: the original problem statement, with the standard instruction \texttt{Please reason step by step, and put your final answer within \textbackslash boxed\{\}}.
    \item \texttt{assistant}: the candidate solution emitted by the policy.
    \item \texttt{user}: the verification instruction shown below, which asks the verifier to search the existing reasoning for concrete mathematical errors and emit a final-line trailer of the form \texttt{Is the existing reasoning correct (Yes/No)? X}.
\end{enumerate}
The instruction explicitly tells the verifier not to demand proof completeness and to output \texttt{No} only when it finds a concrete answer-changing mathematical error — a deliberate design choice that makes the verifier behave as a falsifier rather than a certifier.

\begin{tcolorbox}[title=Prompt Template for the LLM Self-Verifier,fonttitle=\bfseries,breakable,enhanced,
boxrule=0.8mm,
segmentation style={solid,draw=black,line width=0.8mm},
colback=white,
colframe=black,
colbacktitle=white,
coltitle=black,
pad at break=0mm]
\begin{Verbatim}[
  commandchars=\\\{\},
  fontsize=\small,
  breaklines,
  breakindent=0pt,
  breaksymbolleft={},
  breaksymbolright={}
]
[User]
{\color{blue}{\{QUESTION\}}}

[Assistant]
{\color{blue}{\{CANDIDATE\_REASONING\_OR\_SOLUTION\}}}

[User]
You are a skeptical but fair math verifier. Audit the student's solution for concrete mathematical errors, not for proof completeness.

Look for:
- arithmetic mistakes,
- invalid algebraic transformations,
- misuse of a formula or theorem,
- omitted cases or ignored constraints only when you can name a concrete alternative case that changes the final answer,
- a final answer that contradicts the preceding work or relies on a demonstrably false step.

Do not mark a solution wrong for style, brevity, missing explanatory prose, incomplete rigor, or failure to prove optimality, unless you can identify a concrete counterexample or a specific answer-changing mathematical error.

Verdict policy:
- If you find at least one specific answer-changing error, judge No.
- If the final answer is correct and the computations used are correct, judge Yes even if the proof is incomplete.
- If a suspected issue is ambiguous, judge Yes rather than hallucinating an error.

At the end of your response, on a new line, write exactly:
"Is the existing reasoning correct (Yes/No)? X"
where X is either Yes or No.
\end{Verbatim}
\end{tcolorbox}